\pdfoutput=1

\documentclass[11pt]{article}
\usepackage{acl}
\usepackage{times}
\usepackage{latexsym}
\usepackage[T1]{fontenc}
\usepackage[utf8]{inputenc}
\usepackage{microtype}
\usepackage{url}
\usepackage{graphicx}
\usepackage{tabularx}
\usepackage{xcolor}
\usepackage{booktabs}
\usepackage{xspace}
\usepackage{amsmath}
\usepackage{amsfonts}
\usepackage{xfp}
\usepackage{multirow}
\usepackage{subfig}
\usepackage{scalefnt}
\usepackage{stackengine}
\usepackage{hyperref}

\setstackgap{L}{.7\normalbaselineskip}
\stackMath

\newcommand{\emotionname}[1]{\textit{#1}}
\newcommand{\fear}{\emotionname{fear}\xspace}
\newcommand{\joy}{\emotionname{joy}\xspace}
\newcommand{\anger}{\emotionname{anger}\xspace}

\newcommand{\guilt}{\emotionname{guilt}\xspace}
\newcommand{\shame}{\emotionname{shame}\xspace}

\newcommand{\sadness}{\emotionname{sadness}\xspace}

\newcommand{\disgust}{\emotionname{disgust}\xspace}

\newcommand{\other}{\emotionname{other}\xspace}
\newcommand{\noemotion}{\emotionname{no emotion}\xspace}

\newcommand{\otherResp}{\emotionname{other responsibility}\xspace}

\newcommand{\selfResp}{\emotionname{self responsibility}\xspace}

\newcommand{\selfControl}{\emotionname{self control}\xspace}
\newcommand{\situationalControl}{\emotionname{situational control}\xspace}
\newcommand{\situationalResp}{\emotionname{situational responsibility}\xspace}
\newcommand{\extCheck}{\emotionname{external check}\xspace}

\newcommand{\otherControl}{\emotionname{other control}\xspace}

\newcommand{\corpusname}{x-en\textsc{Vent}\xspace}
\newcommand{\F}{F$_1$\xspace}
\newcommand{\expopen}{$\langle{}\text{exp}\rangle$}
\newcommand{\expclose}{$\langle{}\text{/exp}\rangle$}

\title{Experiencer-Specific Emotion and Appraisal Prediction}

\author{%
  Maximilian Wegge, Enrica Troiano, Laura Oberl\"ander \and Roman Klinger \\
  Institut f\"ur Maschinelle Sprachverarbeitung, University of Stuttgart \\
  \texttt{\{firstname.lastname\}@ims.uni-stuttgart.de}
}

\begin{document}
\maketitle
\begin{abstract}
  Emotion classification in NLP assigns emotions to texts, such as
  sentences or paragraphs. With texts like ``I felt guilty when he
  cried'', focusing on the sentence level disregards the standpoint of
  each participant in the situation: the writer (``I'') and the other
  entity (``he'') could in fact have different affective states. The
  emotions of different entities have been considered only partially
  in emotion semantic role labeling, a task that relates semantic
  roles to emotion cue words.  Proposing a related task, we narrow the
  focus on the experiencers of events, and assign an emotion (if any
  holds) to each of them.  To this end, we represent each emotion both
  categorically and with appraisal variables, as a psychological
  access to explaining why a person develops a particular emotion.  On
  an event description corpus, our experiencer-aware models of
  emotions and appraisals outperform the experiencer-agnostic
  baselines, showing that disregarding event participants is an
  oversimplification for the emotion detection task.
\end{abstract}
\section{Introduction} \label{sec:intro}
Computational emotion analysis from text includes various subtasks, with the
most prominent one being emotion classification or regression. Its goal is to
assign an emotion representation to textual units, and the way this is done
typically depends on the domain of the data, the practical application of the
task, and the psychological theories of reference: emotions can be modelled as
discrete labels, in line with theories of basic emotions
\citep{Ekman1992,Plutchik2001}, as valence--arousal value pairs that define an
affect vector space where to situate emotion concepts \citep[illustrated, e.g.,
by ][]{Posner2005}, or as appraisal spaces that correspond to the cognitive
evaluative dimensions underlying emotions\footnote{They are similar 
to a valence--arousal space, but the dimensions
correspond to evaluations of events (i.e., appraisals) that
underlie a certain emotion.} \citep{Scherer2005,Smith1985}.
Irrespective of the adopted representations, most work in the field
detects emotions from a single perspective -- either to recover the
emotion that the writer of a text likely expressed (e.g., with respect
to emotion categories and intensities \citep{Mohammad2018}, and
cognitive categories \cite{Hofmann2020}), or to predict the emotion
that the text elicits in the readers \citep[e.g., using news
articles,][]{Strapparava2007,Bostan2020}. Only a few approaches
combine or compare the reader's with the writer's perspective
\citep[i.a.]{buechel-hahn-2017-readers}. However, none of them looks
at the perspectives of the \textit{participants} in events (both
mentioned or implicit) as described by a text.

Focusing on such perspectives separately is essential to develop an
all-round account of the affective implications that events have. It
would emphasize how the facts depicted in text are amenable to
different ``emotion narratives'', by pushing one or the other
perspective in the foreground. For instance, a possible interpretation
for the sentence ``As the waiter yelled at her, the expression on my
mother's face made all the staff look repulsed'', could be: ``my
mother''$\rightarrow$sadness, ``the waiter''$\rightarrow$anger, and
``the staff''$\rightarrow$disgust. There, one entity is responsible for
an event (screaming), one is influenced by it, and the third is
affected by the emotion emerging in the other (the facial expression,
which can be seen as an event in itself).

Our goal is close to emotion role labeling, a special case of semantic
role labeling (SRL) \cite{Mohammad2018,Kim2018}.  SRL addresses the
question ``Who did What to Whom, Where, When, and How?''
\citep{gildea2000}, emotion SRL asks ``Who feels what, why, and
towards whom?'' \citep{Kim2018}, mainly to detect causes of
emotion-eliciting events \cite{Ghazi2015} for certain entities. Here,
we tackle a variation of this question, namely, ``Who feels what and
under which circumstances?''. The circumstances refer to the
explanation provided by appraisal interpretations, another novelty
that we contribute to the emotion SRL panorama.  Appraisal-based
emotion representations capture entity-specific aspects that lead to
an emotion, as they describe the subjective qualities that an
individual sees in events.

We propose a method for experiencer-specific emotion and appraisal
analysis that bridges emotion classification and semantic role
labeling. Given texts that describe events and that include
annotations for all participants, we assign an emotion and an
appraisal vector to each potential emoter. Our proposal is
computationally simpler than creating a full graph of relations
between causes and entities, as is normally done in (emotion) SRL.
Yet, its fine-grained focus on event participants is beneficial over
traditional classification- and regression-based approaches: by
predicting an emotion and scoring multiple appraisals for each entity,
our model strongly outperforms text-level baselines. Thus, the results
demonstrate that assigning one emotion to the entire instance, or
multiple emotions without considering for whom they hold, is a
simplification of the emotional import of the text.

\section{Related Work}
\label{sec:relatedwork}
In natural language processing, emotions are usually represented as
discrete names following theories of basic emotions
\citep{Ekman1992,Plutchik2001}, or as values of valence and
arousal \citep{russell1977evidence}. Computational models based on
such representations have been applied to many text sources,
including Reddit comments \cite{demszky-etal-2020-goemotions} and tales
\cite{Ovesdotter2005}, 
but also to resources created as part of psychological research. An
example is the ISEAR corpus. It consists of short reports collected in
lab \citep{Scherer1997}, instructing participants to describe events
that caused in them a certain emotion. A similar collection practice
was adopted by \citet{Troiano2019}.  In their enISEAR, crowdworkers
completed sentences like ``I felt [\textsc{emotion name}] when
\ldots'' for seven emotion names.

The emotions of entities are considered in emotion SRL, whose goals
comprise the recognition of emotion cue words, emotion
experiencers/emoters and descriptions of emotion causes and targets
\citep[i.a.]{Mohammad2018,Bostan2020,Kim2018,Campagnano2022}.  Yet,
most work focused on detecting causes (i.e., emotion-triggering
events), and less on other semantic roles
\citep[i.a.]{russo-etal-2011-emocause,chen-etal-2018-joint,chen-etal-2010-emotion,Cheng2017}.

The gap between entity-specific emotion analysis and emotion SRL was
partially filled in by \citet{Troiano2022}. They aimed at better
understanding the readers' attempts to interpret the experience of the
texts' authors. They post-annotated instances from enISEAR with
emotions and 22 appraisal concepts, both for the writer and all other
event participants mentioned in the text.  The appraisal variables
include evaluations of events, as they were likely conducted by the
event experiencers, including if authors felt responsible, if they
needed to pay attention to the environment, whether they found
themselves in control of the situation, and its pleasantness (see
Table 1 in their paper for explanations of the variables). However,
their work was limited to corpus creation and analysis, and did not
provide any modeling of appraisals or emotions in an emotion
experiencer-specific manner. Therefore, it is not clear whether a
simplifying assumption that all entities experience the same emotion
or an actual entity-specific model performs practically better. We
address this concern and show that experiencer-specific modeling is
beneficial.

Finally, our work is related to structured sentiment analysis
\citep{Barnes2021}, in which opinion targets, their polarity, but also
an opinion-holding (or expressing) entity is to be detected. Most studies
focused on sentiment targets and aspects \citep{Brauwers20121}, but
there are also some that aim at detecting the opinion holder
\citep[i.a.]{Kim2006,Wiegand2011,Seki2007,Wiegand2012}.

\begin{table}
  \centering\small
  \begin{tabularx}{\columnwidth}{Xrr}
    \toprule
    Emotion Class & \# inst. & \# exp.\\
    \cmidrule(r){1-1}\cmidrule(l){2-3}
    \textbf{anger} & 259 & 336\\
    \textbf{disgust} & 73 & 87\\
    \textbf{fear} & 173 & 220\\
    \textbf{joy}, pride, contentment & 181 & 265\\
    \textbf{no emotion} & 223 & 269\\
    \textbf{other}, anticipation, hope, & & \\
    surprise, trust & 102 & 117\\
    \textbf{sadness}, disappointment, & & \\
    frustration & 320 & 423\\
    \textbf{shame}, guilt & 282 & 325\\
    \cmidrule(r){1-1}\cmidrule(l){2-3}
    total & 720 & 1329\\
    \bottomrule
  \end{tabularx}
  \caption{Number of instances and experiencer spans annotated for
    each emotion. Non-bold emotion names are concepts in the \corpusname
    data that we merge with bold emotion names in our experiments.}
  \label{tab:data}
\end{table}

\section{Methods and Experimental Setting}
\label{sec:methods}
\paragraph{Model.} We model the task of experiencer-specific emotion
analysis as a classification of instances which consist of
experiencers $e$ in the context of a text
$\mathbf{t}_e=(t_1,\ldots,t_n)$. There can be multiple experiencers in
one text, therefore $\mathbf{t}_e=\mathbf{t}_{e'}$ is possible. Each
experiencer consists of a corresponding token sequence
$(t_i,\ldots,t_j)$ ($1\leq i,j,\leq|\mathbf{t}_e|$), a set of emotion
labels $E_e\in\{\mathrm{anger}, \mathrm{fear}, \mathrm{joy},\ldots\}$,
and a 22-dimensional appraisal vector $\mathbf{a}_e\in[1;5]^{22}$.

To predict $\mathbf{a}_e$ and $E_e$ for each experiencer $e$ with the
help of $\mathbf{t}_e$, we use as input a positional
indicator-encoding of the experiencers in context \citep[inspired by
][]{Zhou2016}. The writer is encoded with an additional special token
$t_o = \mathtt{WRITER}$. We refer to this experiencer-specific model
as \textsc{Exp}.

\begin{table}
  \centering\small
  \begin{tabularx}{\linewidth}{lXp{10mm}c}
    \toprule
    &&\multicolumn{2}{c}{Annotation}\\
    \cmidrule(l){3-4}
    Model & Input instance & Emotion & Appraisal \\
    \cmidrule(r){1-1}\cmidrule(rl){2-2}\cmidrule(rl){3-3}\cmidrule(l){4-4}
    \multirow{2}{*}{\textsc{Exp}}
    &
      \sf\scalefont{0.8}\expopen{}WRITER\expclose{} I felt bad \ldots{} for him
          &
            $\{\text{guilt}\}$ & $(5,1,1,\ldots)$
    \\
    &
      \sf\scalefont{0.8}WRITER I felt bad \ldots for \expopen{}him\expclose{}
          &
            $\{\text{sadness}\}$ & $(1,3,1,\ldots)$
    \\
    \cmidrule(r){1-1}\cmidrule(rl){2-2}\cmidrule(rl){3-3}\cmidrule(l){4-4}
    \textsc{Text} & \sf\scalefont{0.8}WRITER I felt bad \ldots{} for him
          &
            $\{\text{guilt},$ $\text{sadness}\}$ & $(3,2,1,\ldots)$
    \\
    \bottomrule
  \end{tabularx}
  \caption{Example representation at training time for the
    \textsc{Exp} model and the \textsc{Text} baseline for the instance
    ``\underline{WRITER} I felt bad for not being there for
    \underline{him}''.}
  \label{fig:example}
\end{table}

\paragraph{Baseline.} We compare this model to a baseline in which we
simplify the experiencer-specific classification as
text-level classification. During training, we assign the text
$\mathbf{t}$ the union of all emotion labels of all contained
experiencers, namely
$E_{\textbf{t}}=\bigcup_{e, \mathbf{t}_e=\mathbf{t}}E_e$. Analogously,
the aggregation of the appraisal vectors is the centroid of all
experiencers in one text:
$\mathbf{a}_{\textbf{t}}=\frac{1}{|\{e|\mathbf{t}_e=\mathbf{t}\}|}\sum_{e,
  \mathbf{t}_e=\mathbf{t}}\mathbf{a}_e$. We refer to this baseline
model as \textsc{Text}(-based prediction). Table~\ref{fig:example}
examplifies the input representations.

\paragraph{Data Preparation.} We use the x-enVENT data set
\cite{Troiano2022} for our experiments. It consists of 720 event
descriptions, mainly from the enISEAR corpus \cite{Troiano2019}, which
we split into 612 instances for training and 108 instances for testing
(stratified). Each text has been annotated by four annotators and
adjudicated to span-based experiencer annotations with a multi-label
emotion classification and an appraisal vector. We merge infrequent
emotion classes from the original corpus. Table~\ref{tab:data} shows
the label distribution.

\paragraph{Implementation.} We fine-tune Distil-RoBERTa
\cite{Liu2019Roberta} based on the Hugging Face implementation
\cite{Wolf2020}.  For both the emotion classification and the
appraisal regression tasks, we follow a multi-task learning
scheme. All emotion categories are predicted jointly by one model with
a multi-output classification head, analogously with a regression head
for the appraisal vector. The appendix contains implementation
details.\footnote{Our implementation is available at
  \url{https://www.ims.uni-stuttgart.de/data/appraisalemotion}.}

\paragraph{Evaluation.} We evaluate performance by calculating
experiencer-specific \F scores for emotion classification and
Spearman's $\rho$ for appraisal regression.  In the \textsc{Text}
baseline, we project the decision for the text to each experiencer
that it contains.

\begin{table}[t]
  \centering\small
  \setlength{\tabcolsep}{6pt}
  \begin{tabular}{l rrr rrr r}
    \toprule
    & \multicolumn{3}{c}{\textsc{Text}} & \multicolumn{3}{c}{\textsc{Exp}}&\\
    \cmidrule(r){2-4}
    \cmidrule(r){5-7}
    Emotion Class
    & P & R & \F & P & R & \F & $\Delta$\F \\
    \cmidrule(r){1-1}\cmidrule(rl){2-4}\cmidrule(rl){5-7}\cmidrule(l){8-8}
    anger      & 40 & 82 & 54 & 60 & 80 & 68 & $+14$ \\
    disgust    & 50 & 93 & 65 & 60 & 80 & 69 & $+4$  \\
    fear       & 44 & 86 & 58 & 53 & 71 & 61 & $+3$  \\
    joy        & 55 & 70 & 62 & 61 & 77 & 68 & $+6$ \\
    no emotion & 29 & 80 & 42 & 51 & 80 & 62 & $+20$ \\
    other      & 11 & 10 & 10 & 14 & 10  & 12  & $+2$ \\
    sadness    & 47 & 90 & 62 & 62 & 93 & 74 & $+12$ \\
    shame      & 34 & 89 & 49 & 48 & 85 & 61 & $+12$ \\
    \cmidrule(r){1-1}\cmidrule(rl){2-4}\cmidrule(rl){5-7}\cmidrule(l){8-8}
    Macro avg. & 39 & 75 & 51 & 51 & 72 & 60 & $+9$  \\
    Micro avg. & 40 & 79 & 53 & 55 & 78 & 64 & $+11$ \\
    \bottomrule
  \end{tabular}
  \caption{Emotion classification results of the \textsc{Text}-based
    baseline which is not informed about experiencer-specific emotions
    with our emotion experiencer-specific model \textsc{Exp}.}
  \label{tab:emoresults}
\end{table}

\section{Results}
\label{sec:results}
\paragraph{Quantitative Evaluation.} Tables~\ref{tab:emoresults}
and~\ref{tab:appraisalresults} show the results.  For
emotion classification, we report precision, recall, and \F
measures for the baseline \textsc{Text} and the experiencer-specific
predictions by \textsc{Exp} in Table~\ref{tab:emoresults}.
\textsc{Exp} substantially outperforms \textsc{Text} in terms of \F
score.  This trend holds across all emotion categories, as a result of
an increased precision, which is intuitively reasonable, because the
\textsc{Exp} model learns to distribute the emotions that are
contained in a text to individual experiencers, while the
\textsc{Text} baseline distributes all emotions to all experiencers
equally, leading to an increased recall. The most substantial
improvements are observed for \anger (+14), \sadness (+12) and \shame
(+12) as well as for \noemotion (+20).  These results are in line with
the corpus analysis by \citet{Troiano2022}. They found that some
emotions are often shared between different experiencers within one
text, but others occur in common pairs, namely \guilt--\anger,
\noemotion--\sadness, \guilt--\sadness and \shame--\anger. Noteworthy
is the category \noemotion, which commonly occurs with all other
emotions \citep[Figure 4]{Troiano2022}.
\begin{table}[t]
  \centering\small
  \begin{tabular}{lccr}
    \toprule
    & \textsc{Text} & \textsc{Exp} & \\
    \cmidrule(r){2-2}\cmidrule(r){3-3}
    Appraisal Dimension & $\rho$ & $\rho$&$\Delta\rho$\\
    \cmidrule(r){1-1}\cmidrule(rl){2-2}\cmidrule(rl){3-3}\cmidrule(l){4-4}
    Suddenness           & 0.32 & 0.54 & $+0.22$\\
    Familiarity          & 0.17 & 0.37 & $+0.20$\\
    Pleasantness         & 0.34 & 0.60 & $+0.26$\\
    Understand           & 0.24 & 0.30 & $+0.06$\\
    Goal relevance       & 0.15 & 0.33 & $+0.18$\\
    Self responsibility   & 0.31 & 0.68 & $+0.37$\\
    Other responsibility & 0.33 & 0.68 & $+0.35$\\
    Situational respons. & 0.59 & 0.68 & $+0.09$\\
    Effort               & 0.33 & 0.54 & $+0.21$\\
    Exert                & 0.97 & 0.25 & $-0.72$\\
    Attend               & 0.27 & 0.41 & $+0.14$\\
    Consider             & 0.55 & 0.62 & $+0.07$\\
    Outcome probability  & 0.14 & 0.38 & $+0.24$\\
    Expect. discrepancy  & 0.43 & 0.54 & $+0.11$\\
    Goal conduciveness   & 0.47 & 0.65 & $+0.18$\\
    Urgency              & 0.20 & 0.25 & $+0.05$\\
    Self control         & 0.36 & 0.64 & $+0.28$\\
    Other control        & 0.41 & 0.69 & $+0.28$\\
    Situational control  & 0.63 & 0.67 & $+0.04$\\
    Adjustment check     & 0.39 & 0.56 & $+0.17$\\
    Internal check       & 0.47 & 0.58 & $+0.11$\\
    External check       & 0.66 & 0.54 & $-0.12$\\
    \cmidrule(r){1-1}\cmidrule(rl){2-2}\cmidrule(rl){3-3}\cmidrule(l){4-4}
    Avg.                 & 0.44 & 0.54 & $+0.09$\\
    \bottomrule
  \end{tabular}
  \caption{Appraisal regression results of the \textsc{Text}-based
    baseline and the experiencer-specific model \textsc{Exp}. The
    average has been calculated via FisherZ-Transformation.}
  \label{tab:appraisalresults}
\end{table}
The performance increase for \joy, \fear and \disgust is less
distinct: these emotions are likely shared by all event
experiencers.

For the appraisal predictions, we report Spearman's $\rho$ in
Table~\ref{tab:appraisalresults}.  We observe an improved performance
prediction across nearly all dimensions. Appraisals
that distinguish between who caused the event and who had the power to
influence it (\textit{self} vs.\ \textit{other}) show the most
substantial improvement, namely \selfResp (+0.37) and \selfControl
(+0.28), as well as \otherResp (+0.35) and \otherControl (+0.28).
This is reasonable -- the \textit{self} and \textit{other} are often
mutually exclusive. This interaction of appraisals cannot be exploited
by purely text-level prediction models.  However, if an event is
caused by external factors, like \situationalResp (+.09) and
\situationalControl (+.04), all experiencers are equally affected by
it. The decrease in performance for \extCheck ($-$0.12) might be
explained by the fact that this dimension is often shared between
experiencers, rendering the \textsc{Text} model sufficiently
efficient.

\begin{table*}[t]
  \centering\small
  \subfloat[][Example Texts\label{sf1}]{%
    \begin{tabularx}{1.0\linewidth}{lX}
      \toprule
      ID & Text \\
      \cmidrule(r){1-1}\cmidrule(l){2-2}
      1 & I felt \dots working in the street seeing faeces of dogs. \underline{The
          owners} should take care of them but are being so lazy and
          neglected, that is terrible.
      \\
      2 & I felt \dots when I remember being part of \underline{a
          group of children} at school who verbally bullied
          \underline{another child}.
      \\
      3 &  I felt \dots when I lost \underline{my sister}'s necklace that I had
          borrowed.
      \\
      4 & I felt \dots when \underline{my ex husband} was hateful towards \underline{our
          children}.
      \\
      5 & I felt \dots when \underline{my son} was born.
      \\
      \bottomrule
    \end{tabularx}    
  }

  \newcommand{\xxx}{0}
  \newcommand{\xc}[1]{%
    \renewcommand{\xxx}{#1}%
    \setlength{\fboxsep}{0pt}%
    \fbox{\textcolor{black!\fpeval{(\xxx-1)*100/4}!white}{\rule{0.9em}{0.9em}}}%
  }
  \newcommand{\xcc}[4]{%
    \xc{#1}\xc{#2}\xc{#3}\xc{#4}
  }
  
  \subfloat[][Annotations \label{sf2}]{%
    \begin{tabular}{llllllllll}
      \toprule
      && \multicolumn{2}{c}{Gold} & \multicolumn{2}{c}{\textsc{Text}} & \multicolumn{2}{c}{\textsc{Exp}} \\
      \cmidrule(lr){3-4}\cmidrule(lr){5-6}\cmidrule(l){7-8}
      ID & Experiencer Text & Emotion & Appraisal & Emotion & Appraisal & Emotion & Appraisal \\
      \cmidrule(r){1-1}\cmidrule(rl){2-2}\cmidrule(rl){3-3}\cmidrule(rl){4-4}\cmidrule(rl){5-5}\cmidrule(rl){6-6}\cmidrule(rl){7-7}\cmidrule(rl){8-8}
      1 & Writer & a d & \xcc{1.0}{5.0}{1.5}{4.3} & a d no sa & \xcc{2.5}{2.6}{2.4}{2.5} & a d sa& \xcc{2.0}{2.5}{2.1}{2.2}  \\
      & The owners & no & \xcc{4.3}{1.0}{3.5}{1.5} & a d no sa & \xcc{2.5}{2.6}{2.4}{2.5} & no & \xcc{3.1}{2.3}{3.1}{2.1} \\
      \cmidrule(r){1-1}\cmidrule(rl){2-2}\cmidrule(rl){3-3}\cmidrule(rl){4-4}\cmidrule(rl){5-5}\cmidrule(rl){6-6}\cmidrule(rl){7-7}\cmidrule(rl){8-8}
      2 & Writer & sh & \xcc{4.7}{4.0}{4.3}{2.3} & a no sa sh & \xcc{2.8}{3.0}{2.6}{2.8} & sh & \xcc{2.1}{4.0}{2.3}{3.5} \\
      & a group of children & j sh & \xcc{5.0}{1.0}{4.7}{1.3} & a no sa sh & \xcc{2.8}{3.0}{2.6}{2.8} & a j no sh & \xcc{3.3}{3.3}{3.3}{2.9} \\
      & another child & sa & \xcc{1.0}{5.0}{1.5}{4.8} & a no sa sh & \xcc{2.8}{3.0}{2.6}{2.8} & a f sa & \xcc{1.4}{4.3}{1.9}{3.9} \\
\cmidrule(r){1-1}\cmidrule(rl){2-2}\cmidrule(rl){3-3}\cmidrule(rl){4-4}\cmidrule(rl){5-5}\cmidrule(rl){6-6}\cmidrule(rl){7-7}\cmidrule(rl){8-8}
      3 & Writer & sa sh & \xcc{4.8}{1.0}{3.0}{1.5} & sa sh & \xcc{2.6}{1.9}{2.2}{1.7} & sa sh & \xcc{3.1}{1.7}{2.4}{1.4} \\
      & my sister & sa no & \xcc{1.0}{4.7}{1.7}{3.7} & sa sh & \xcc{2.6}{1.9}{2.2}{1.7} & sa no & \xcc{1.8}{3.0}{1.7}{2.8} \\
\cmidrule(r){1-1}\cmidrule(rl){2-2}\cmidrule(rl){3-3}\cmidrule(rl){4-4}\cmidrule(rl){5-5}\cmidrule(rl){6-6}\cmidrule(rl){7-7}\cmidrule(rl){8-8}
      4 & Writer & a sa & \xcc{1.0}{5.0}{1.5}{4.5} & a f j no sa sh & \xcc{2.7}{3.1}{2.6}{2.9} & a sa & \xcc{1.1}{4.2}{1.6}{4.0} \\
      & my ex husband & a sh & \xcc{5.0}{1.5}{4.5}{1.8} & a f j no sa sh & \xcc{2.7}{3.1}{2.6}{2.9} & a j sh & \xcc{4.0}{2.0}{3.6}{1.9}\\
      & our children & sa & \xcc{1.3}{5.0}{1.8}{4.5} & a f j no sa sh & \xcc{2.7}{3.1}{2.6}{2.9} & a f sa & \xcc{1.3}{4.3}{1.8}{4.0} \\
\cmidrule(r){1-1}\cmidrule(rl){2-2}\cmidrule(rl){3-3}\cmidrule(rl){4-4}\cmidrule(rl){5-5}\cmidrule(rl){6-6}\cmidrule(rl){7-7}\cmidrule(rl){8-8}
      5 & Writer & j & \xcc{4.5}{3.5}{2.3}{3.3} & j o no & \xcc{2.2}{1.6}{2.2}{1.5} & j & \xcc{3.0}{2.6}{2.6}{2.6} \\
      & my son & no & \xcc{1.0}{5.0}{1.0}{3.0} & j o no & \xcc{2.2}{1.6}{2.2}{1.5} & j no & \xcc{1.7}{1.5}{1.2}{1.6} \\
      \bottomrule
    \end{tabular}
  }
  \caption{Examples of \textsc{Exp} and \textsc{Text}
    predictions. a: anger, d: disgust, no: no emotion, o:other, sa: sadness, sh:
    shame, f: fear, j: joy.  The boxes show the appraisal
    \selfResp, \otherResp, \selfControl,
    \otherControl, with values between \protect\xc{1} and \protect\xc{5}.}
  \label{tab:examples}
\end{table*}
\paragraph{Analysis.}
We show some examples in Table~\ref{tab:examples} that highlight the
usefulness of \textsc{Exp} over \textsc{Text}. Next to the emotion
classification annotations and predictions from both models, we show
the appraisals of \selfResp/\otherResp and \selfControl/\otherControl.
In each example, the writer is one emotion experiencer. All other
experiencers are underlined.

We observe that the \textsc{Text} model has a tendency to
predict the union of the emotions for all experiencers, but sometimes
predicts more additional categories. This is a consequence of the
tendency towards high recall predictions of this model.
In Example 1, both \textsc{Exp} and \textsc{Text} correctly assign the
emotions \anger, \disgust and \noemotion, but only \textsc{Exp}
distributes them correctly between ``Writer'' and ``The owners''
(\sadness is wrongly detected by both models).
In Example 2, \joy is not predicted by \textsc{Text}, but correctly
assigned to ``a group of children'' by \textsc{Exp}. \textsc{Exp}
further distributes \shame and \sadness to the correct entities (with
a mistake assigning \anger and \noemotion to ``a group of children''
as well as \anger and \fear to ``another child'').
In Example 3, \textsc{Exp} correctly assigns \sadness and \shame to
``Writer'' and \sadness and \noemotion to ``my sister'', while
\textsc{Text} fails to detect \noemotion.
In Example 4, \textsc{Exp}'s prediction of \anger and \fear (for ``our
children'') could be accepted to be correct despite it not being in
line with the gold annotation. \textsc{Exp} further predicts the
correct emotions for ``Writer'' (but makes a mistake assigning \joy to
``my ex husband'').
In Example 5, the emotions of ``Writer'' are correctly assigned; ``my
son'' is wrongly assigned \joy in addition to \noemotion
(\textsc{Text} mistakenly predicts \other as well).  However, the
correctness of this annotation is debatable.

Maximal values for the gold appraisal values for self/other control
and self/other responsibility are, in nearly all cases, mutually
exclusive across experiencers. The \textsc{Text} model is not informed
about that and distributes the values across all entities. The
\textsc{Exp} model does indeed recover the individual values for the
appraisals, but to varying degrees.
In Examples 2, 3, and 4, nearly all experiencers receive appraisal
values close to the gold annotations. Example 2 appears to be
challenging: the writer has a high gold annotation value for \selfResp
which is not automatically detected. Further, ``a group of children''
receives the same values for the four appraisals.
Examples 1/5 are cases in which the appraisal prediction does not
work as expected.

\section{Discussion and Conclusion}
We presented the first approach of experiencer-specific emotion
classification and appraisal regression. Our evaluation on event
descriptions shows the need for such methods, and that a text-instance
level annotation is a simplification.

This work provides the foundation for future research focused on texts
in which multiple emotion labels co-occur, including reader/writer
combinations or turn-taking dialogues. We propose to integrate
experiencer-specific emotion modeling within such settings, for
instance in novels, or news articles. It can also enrich the work of
emotion recognition in dialogues \citep{Poria2019}: Chains of
emotions have been modeled, but not considering mentioned entities.

Our work focused on a corpus that has been annotated specifically for
writers' and entities' emotions. There exist, however, also other
corpora with experiencer-specific emotion annotations, namely emotion
role labeling resources
\citep{Kim2018,Bostan2020,Campagnano2022,Mohammad2014}. In addition to
other information, they also provide experiencer-specific emotion
labels, though not in such an event-focused context. Still, modeling
them following our method needs to be compared to more traditional
approaches that aim at recovering the full role labeling graph.

Our approach to encoding the experiencer position in the classifier has
been a straightforward choice. Other model architectures
\citep[including positional embeddings,][]{Wang2020} might perform
better. Another interesting methodological avenue is to model the
predictions of multiple experiencers jointly to exploit their
relations.

Finally, an open question is how to incorporate information from
existing resources that are not labeled with experiencer-specific
information. For instance, \citet{Troiano2022b} provide appraisal and
emotion annotations for many more instances that might be beneficial
in a transfer-learning setup.

\section*{Acknowledgements}
This research is funded by the German Research Council (DFG), project
``Computational Event Analysis based on Appraisal Theories for Emotion
Analysis'' (CEAT, project number KL 2869/1-2).

\bibliography{lit}

\appendix

\section{Implementation Details.} We fine-tune Distil-RoBERTa
\cite{Liu2019Roberta} as implemented in the Hugging Face
library\footnote{\url{https://huggingface.co/distilroberta-base}}
\cite{Wolf2020} and leave default parameters unchanged.  For both the
emotion classification and the appraisal regression tasks, we follow a
multi-task learning scheme. All emotion categories are predicted
jointly by one model with a multi-output classification head,
analogously with a regression head for the appraisal vector
prediction.  The classification head consists of a linear layer with
dropout (0.5) and ReLU activation function, followed by a final linear
layer with sigmoid activation.  For the appraisal regression, the
sigmoid activation function in the final layer is replaced by a linear
activation. We use binary cross entropy loss in the emotion classifier
and mean squared error loss in the appraisal regressor.  Both models
are trained for 10 epochs without early stopping.  We use the Adam
optimizer \citep{adam} with weight decay (0.001) and a learning rate
of $2\cdot10^{-5}$.  The weights of each layer are initialized using
the Xavier uniform initialization \citep{xavier}.  The hyperparameters
and architecture have been decided on via 10-fold cross validation on
the training data.

\end{document}